\documentclass[conference]{IEEEtran}
\IEEEoverridecommandlockouts
\usepackage{cite}
\usepackage{amsmath,amssymb,amsfonts}
\usepackage{algorithmic}
\usepackage{graphicx}
\usepackage{textcomp}
\usepackage{xcolor}
\def\BibTeX{{\rm B\kern-.05em{\sc i\kern-.025em b}\kern-.08em
    T\kern-.1667em\lower.7ex\hbox{E}\kern-.125emX}}

\usepackage{overpic}
\usepackage{hyperref}
\usepackage{makecell}

\usepackage{scalerel}
\usepackage{tikz}
\usetikzlibrary{svg.path}

\definecolor{orcidlogocol}{HTML}{A6CE39}
\tikzset{
    orcidlogo/.pic={
        \fill[orcidlogocol] svg{M256,128c0,70.7-57.3,128-128,128C57.3,256,0,198.7,0,128C0,57.3,57.3,0,128,0C198.7,0,256,57.3,256,128z};
        \fill[white] svg{M86.3,186.2H70.9V79.1h15.4v48.4V186.2z}
        svg{M108.9,79.1h41.6c39.6,0,57,28.3,57,53.6c0,27.5-21.5,53.6-56.8,53.6h-41.8V79.1z M124.3,172.4h24.5c34.9,0,42.9-26.5,42.9-39.7c0-21.5-13.7-39.7-43.7-39.7h-23.7V172.4z}
        svg{M88.7,56.8c0,5.5-4.5,10.1-10.1,10.1c-5.6,0-10.1-4.6-10.1-10.1c0-5.6,4.5-10.1,10.1-10.1C84.2,46.7,88.7,51.3,88.7,56.8z};
    }
}

\newcommand\orcidicon[1]{\href{https://orcid.org/#1}{\mbox{\scalerel*{
                \begin{tikzpicture}[yscale=-1,transform shape]
                \pic{orcidlogo};
                \end{tikzpicture}
            }{|}}}}

\begin{document}

\makeatletter
\newcommand{\linebreakand}{%
  \end{@IEEEauthorhalign}
  \hfill\mbox{}\par
  \mbox{}\hfill\begin{@IEEEauthorhalign}
}
\makeatother

\title{DLSIA: Deep Learning for Scientific Image Analysis\\

}

\author{
    \IEEEauthorblockN{Eric J. Roberts\IEEEauthorrefmark{1}\IEEEauthorrefmark{2}$^{\textsuperscript{\orcidicon{0000-0001-8565-1424}}}$,
    Tanny Chavez\IEEEauthorrefmark{3}$^{\textsuperscript{\orcidicon{0000-0001-9317-2896}}}$,
    Alexander Hexemer\IEEEauthorrefmark{1}\IEEEauthorrefmark{2}$^{\textsuperscript{\orcidicon{00000-0002-5269-0125}}}$,
    Petrus H. Zwart\IEEEauthorrefmark{1}\IEEEauthorrefmark{2}\IEEEauthorrefmark{4}$^{\textsuperscript{\orcidicon{0000-0003-3315-4092}}}$}
    
    \IEEEauthorblockA{\IEEEauthorrefmark{1}\textit{Center for Advanced Mathematics for Energy Research Applications}, 
    \textit{Lawrence Berkeley National Laboratory},
    Berkeley, CA, USA}
    
    \IEEEauthorblockA{\IEEEauthorrefmark{2}
    \textit{Molecular Biophysics and Integrated Bioimaging Division, Lawrence Berkeley National Laboratory},Berkeley, CA, USA}
    
   \IEEEauthorblockA{\IEEEauthorrefmark{3}\textit{Advanced Light Source}, 
    \textit{Lawrence Berkeley National Laboratory},
    Berkeley, CA, USA}
   
    \IEEEauthorblockA{\IEEEauthorrefmark{4}
    \textit{Berkeley Synchrotron Infrared Structural Biology Program, Lawrence Berkeley National Laboratory}\\
    }

    \IEEEauthorblockA{Emails: EJRoberts@lbl.gov, TanChavez@lbl.gov, AHexemer@lbl.gov, \textbf{PHZwart@lbl.gov}}
}
\maketitle

\begin{abstract} We introduce DLSIA (\textbf{D}eep \textbf{L}earning for \textbf{S}cientific \textbf{I}mage \textbf{A}nalysis), a Python-based machine learning library that empowers scientists and researchers across diverse scientific domains with a range of customizable convolutional neural network (CNN) architectures for a wide variety of tasks in image analysis to be used in downstream data processing, or for experiment-in-the-loop computing scenarios. DLSIA features easy-to-use architectures such as autoencoders, tunable U-Nets, and parameter-lean mixed-scale dense networks (MSDNets). Additionally, we introduce sparse mixed-scale networks (SMSNets), generated using random graphs and sparse connections. As experimental data continues to grow in scale and complexity, DLSIA provides accessible CNN construction and abstracts CNN complexities, allowing scientists to tailor their machine learning approaches, accelerate discoveries, foster interdisciplinary collaboration, and advance research in scientific image analysis.
\end{abstract}


\section{Introduction}
\subsection{Purpose \& Motivation}
Scientific image analysis forms a crucial component of numerous workflows at user facilities, generating an abundance of datasets that each possess unique characteristics. Given the distinct nature of these datasets, the need frequently arises to craft custom solutions tailored to individual experiments. Convolutional neural networks (CNNs), along with other machine learning tools, prove extremely valuable in this regard, capable of addressing a variety of analysis needs and producing insightful results. The unique aspect of scientific data analysis in such settings often necessitates the creation of bespoke solutions tailored to individual experiments, providing optimal results given the data's specific characteristics. CNNs, along with a host of other machine learning tools, present themselves as exceptionally suitable for such tasks due to their flexibility and the broad array of potential applications they cater to.

%

\subsection{Background \& Prior Art}
Convolutional neural networks (CNNs) have emerged as a transformative class of machine learning models specifically designed to unravel patterns and extract meaningful features from various forms of data. Having gained significant popularity in the scientific community, CNNs are \emph{particularly} well-suited for tackling image analysis tasks, including object detection, image classification, and pixel-by-pixel semantic segmentation. The unique strength of CNNs lies in their ability to autonomously learn discriminative features directly from the data itself, eliminating the need for laborious manual feature engineering. By training on large datasets with labeled examples, CNNs can learn to recognize specific objects, identify anomalies, or detect subtle patterns. Moreover, CNNs remain a versatile tool, allowing researchers from different backgrounds to choose from a variety of different CNN architectures that can denoise, reconstruct, and segment images \cite{xing2017deep, kaur2018review, manifold2019denoising, gong2019machine, kromp2020annotated, jung2014impact}, or perform higher level tasks from among their diverse scientific disciplines, including automated structure and material classification and data-driven discovery in X-ray scattering \cite{kiapour2014materials, liu2019convolutional, deyhle2018automated, douarre2018transfer, wang2017x}, biological \cite{radivojevic2020machine, waldchen2018machine}, crystallographic \cite{ziletti2018insightful, kirman2020machine, sun2019accelerated}, and signal processing \cite{tabar2016novel, schirrmeister2017deep, lawhern2018eegnet, likamwa2016redeye} settings.

Inspired by the structure and hierarchy of visual representations in the animal and human visual system in which each individually specialized cortical areas work in tandem to identify objects in a visual field \cite{lindsay2021convolutional}, CNNs apply small filters across the input data in the form of several adjacently connected convolutional layers to learn relevant features and progressively transform sequences of raw pixel intensities into higher-level representations. This layered computation and hierarchical abstraction empowers CNNs to discern intricate patterns and identify key features in data. Contrasted with earlier, more-traditional fully-connected neural networks (FCNs) \cite{rosenblatt1958perceptron}, CNNs enforce a more-localized learning of image features and require \emph{far} fewer weights to learn, resulting in deeper CNN architectures with more-targeted learning \cite{goodfellow2016deep}.

While the widespread adaptability of CNNs has made them a prevalent tool across various scientific domains, not all scientific researchers possess the expertise or knowledge required to construct and train these networks effectively.  Access to user-friendly libraries with pre-built networks is invaluable for individuals lacking deep understanding of CNNs. These libraries offer a convenient way to deploy CNNs without dealing with network architecture intricacies. Researchers can focus on their domain expertise by leveraging these libraries instead of building CNNs from scratch. The flexibility of these libraries enables iterative experimentation, allowing researchers to easily swap network architectures and adjust hyperparameters to find the best configurations for their problems. Access to state-of-the-art networks saves time and resources, while promoting interdisciplinary collaboration by abstracting the complexities of CNN construction and training, as researchers can focus on their areas of expertise while leveraging the power of CNNs for their analyses. 

In summary, the prevalence of CNNs in the sciences necessitates user-friendly libraries that simplify their construction and training, allowing scientists to stay at the forefront of CNN research without the need for extensive expertise in deep learning. To address these challenges and expedite the process of incorporating machine learning into scientific image analysis workflows, we introduce DLSIA (Deep Learning for Scientific Image Analysis), a Python-based, general-purpose machine learning library offering a flexible and customizable environment for generating custom CNN architectures and an extensive suite of tools designed to empower scientists and researchers from diverse scientific domains, including beamline scientists, biologists, and researchers in X-ray scattering. DLSIA, found here \url{https://dlsia.readthedocs.io/en/latest/}, enables a seamless integration of custom CNN architectures and other advanced machine learning methods into common workflows, providing researchers with the means to rapidly test and implement different analysis approaches within a unified framework, dramatically increasing efficiency and adaptability. Whether the task at hand involves image classification, anomaly detection, or any other complex pattern recognition, DLSIA offers a streamlined, efficient platform that enables users to explore, compare, and customize a wide array of CNN architectures, facilitating a systematic investigation of what works, what doesn't, and what is best suited for their specific scientific problems.

\subsection{The DLSIA software library}

The core focus of DLSIA lies in its ability to bridge the gap between cutting-edge deep learning techniques and the challenges encountered in scientific image analysis. By offering a comprehensive collection of user-customizable CNNs, including autoencoders, tunable U-Nets (TUNet), mixed-scale dense networks (MSDNet), and more novel randomized sparse mixed-scale networks (SMSNet), DLSIA allows researchers to harness the power of state-of-the-art deep learning while tailoring a network architecture to the specific demands of their scientific investigations. This flexibility empowers users to fine-tune CNNs, select appropriate layers, optimize hyperparameters, and explore diverse architectural variations, enabling a comprehensive exploration of the rich design space inherent in deep learning-based image analysis.

DLSIA facilitates seamless integration with various scientific datasets and promotes reproducible research through its intuitive and extensible PyTorch Application Programming Interface (API). It offers a rich set of functionalities for data preprocessing, model training, validation, and evaluation, while also providing convenient visualization tools to aid in the interpretation and analysis of results. With its user-centric design philosophy, DLSIA aims to empower scientists across domains to leverage the potential of CNNs for scientific image analysis, ultimately accelerating discoveries and advancing research in a wide range of scientific fields.

The rest of the manuscript is organized as follows: Sect.~\ref{sect:DLSIA_CNNs} offers an in-depth look at the CNN architectures offered, Sect.~\ref{sect:utility} describes the different utility functions, data loaders, training regiments, uncertainty quantification available to DLSIA users, we validate DLSIA CNN architectures through various applications on experimental data in Sect.~\ref{sect:applications}, and Sect.~\ref{sect:discussion} concludes with a discussion of DLSIA results and viability. 

\section{DLSIA Deep Convolutional Neural Networks} \label{sect:DLSIA_CNNs}

Convolutional neural networks (CNNs) are deep learning models that excel at visual data analysis. In general, CNNs capture features by applying many convolutional filters, or kernels, to local regions of the data via several adjacently connected convolutional layers. The filters are square matrices with adjustable weights that serve as ``windows" observing a specific region of the image. By learning the filters' weights via network training and optimization, CNNs can identify various features within the image. 

We highlight below the different CNN architectures available in the DLSIA software library. Each available network varies in its sequencing of layers and addition of nonlinear activation, pooling, and normalization layers to decompose images into complex hierarchical structures and increase the expressive power. But true to the original goal of DLSIA, all networks are fully customizable with an array of user-specified hyperparameters available to toggle.

\subsection{Tunable U-Nets} \label{sect:tunets}

Included in the DLSIA software suite are tunable variant of U-Nets (TUNets), a popular and effective deep convolutional neural network \cite{ronneberger2015u}. Inspired by autoencoders (Sect.~\ref{sect:autoencoders}) and first introduced for the segmentation of biomedical images, its distinctive U-shaped architecture consists of typically-mirrored contractive encoder and expansive decoder halves. Contextual information and features are captured by the contractual encoder phase, made up of a predefined number of layers $d$. Each individual layer consists of stacked $3\times 3$ unpadded convolutional operators, nonlinear activation (typically in the form of the rectified linear unit (ReLU) function), and batch normalization to expedite the learning process \cite{ioffe2015batch}. Between each layer, max-pooling operations reduce the spatial dimensionality to ease computational costs, introduce translational equivariance \cite{finzi2020generalizing}, and encourage higher-level feature extraction. Next, the upsampling phase of the expansive decoder half, each layer mirrors the stacked convolution-activation-normalization block, while transposed convolutions between layers recover the previously compressed spatial dimensions. This effectively projects the encoder's learned features into the higher resolutions of the original image space to predict a pixel-by-pixel semantic segmentation. In total, the modified upsampling phase offers many more convolutional channels in which the contextual information may be propagated, allowing one to combine and correlate local features of an image with its behavior at larger length scales \cite{noh2019scale, springenberg2014striving}. Moreover, long-reaching skip connections are introduced in the form of channel-wise concatenations of intermediate feature maps between adjacent contractive and expansive phases, allowing for an aggregation of multi-scale feature representation at different network stages \cite{zhou2018unet++, zhou2019unet++, kumar2018u, drozdzal2016importance}.

TUNet performance on different applications relies significantly on the the various hyperparameters that govern the network architecture \cite{kinnison2018shadho, li2021lightweight, berral2021distributing}. As such, the DLSIA API offers full flexibility in creating and deploying TUNets of custom sizes and morphology by allowing the user to define the four following architecture-governing hyperparameters:

\begin{enumerate}
    \item Depth $d$: the number of layers in the TUNet. A depth of $d$ will contain $d$ layers of dual convolutions and accompanying intralayer operations in each of the encoder and decoder phases, with
    $d-1$ mirrored max-pooling, up-convolutions, and concatenation steps between each layer.
    \item Number of initial base channels $c_b$: the input data is mapped to this number of feature channels after the initial convolution.
    \item Growth rate $r$: the growth rate/decay rate of feature channels between successive layers,
    \item Hidden rate $r_h$: the growth rate/decay rate of feature channels within each individual layer, between each layers' successive convolutions.
\end{enumerate}

\noindent Additionally, DLSIA defaults to ReLU nonlinear activation and batch normalization after each convolution operation, though the user is free to apply any activation or normalization consistent with PyTorch syntax. A U-Net schematic of depth $d=4$ is shown in Fig.~\ref{fig:unet} depicting the order of operations and evolution of channels and spatial dimensions along the contracting and expanding halves. We note that the growth and hidden rates of feature channel growth and decay may be non-integers. 

\begin{figure}
    \centering
    \includegraphics[width=0.45\textwidth]{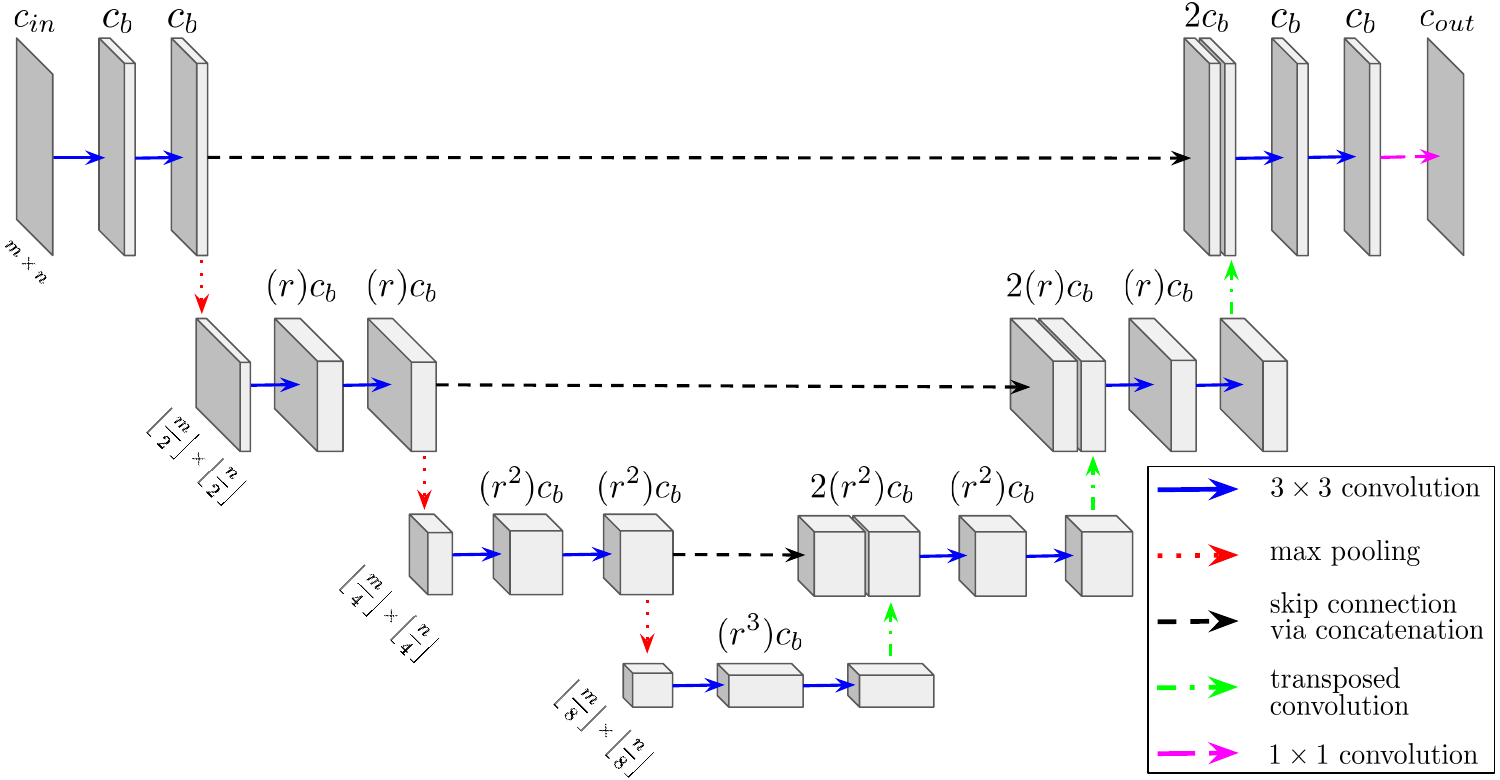}%
    
    \caption{Diagram of a two-dimensional, four-layer tunable U-Net congruent with input data of $c_{in}$ channels and spatial dimensions $m$ and $n$. Among the user-defined hyperparameters on display are the initial base channels $c_b$ and the channel growth factor $r$, both of which control the size of the network, and thus its potential expressive power. The hidden growth rate $r_h$ is set to 1 for simplicity. We note that DLSIA easily accommodates volumetric data by simply replacing all convolutions (and associated layer normalization) with their three-dimensional counterparts.}
    \label{fig:unet}
    
\end{figure}

\subsection{Convolutional Autoencoder} \label{sect:autoencoders}

Convolutional autoencoders are a deep, unsupervised neural network framework generally tasked with learning feature extraction for the purpose of reconstructing the input \cite{rumelhart1985learning, lecun1998gradient}.  While relatively simple in structure and acting as a precursor to U-Net encoder-decoder structure, autoencoders use convolutional layers and max-pooling operations between adjacent layers to exploit the feature extraction properties in the beginning encoder half of its architecture. As shown in Fig.~\ref{fig:autoencoder}, the encoder half terminates at a single-dimensional latent space of features, often referred to as the latent space representation. This informational ``bottleneck" forces the network to learn only the most important features and contextual information. The second half of the network, the decoder, concludes with alternating transposed convolutions and blocks of dual convolutionals to project the information back to the input space and learn the reconstruction of input data

\begin{figure}
    \includegraphics[width=.45\textwidth]{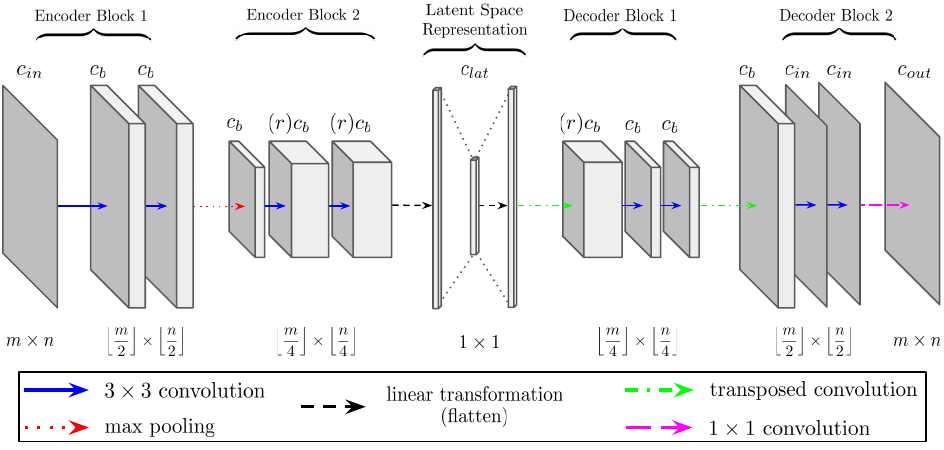}%
        \centering

        \caption{Schematic overview of a two-layer autoencoder congruent with input data of $c_{in}$ channels and spatial dimensions $m$ and $n$. DLSIA provides the flexibility to adjust the following hyperparameters: initial base channels $c_b$, channel growth factor $r$, and length of latent space vector $c_{lat}$.}
        \label{fig:autoencoder}

\end{figure}

DLSIA instantiation of autoencoders once again reflects that of the tunable U-Nets. Users once again may find the autoencoder with the appropriate expressive power to suit their needs by toggling the number of layers $d$, the initial number of base channels $c_b$, and the growth rate $r$ of the convolutional channels. Additionally, users are free to experiment with different sizes of latent space vectors with the  $c_{lat}$ hyperparameter.

\subsection{Mixed-scale Dense Convolutional Neural Networks} \label{sect:msdnets}

The mixed-scale dense network (MSDNet) was developed as a deep learning framework with a relatively simple architecture containing roughly two to three orders of magnitude \emph{fewer} trainable parameters \cite{pelt2018mixed, pelt2018improving} than U-Nets and typical encoder-decoder networks. This reduction in model complexity reduces the risk of overfitting, a common problem in neural networks that occurs when the model fits the training data too closely, resulting in poor generalization \cite{li2019research, ying2019overview}. MSDNets reduce the model complexity in two ways. Firstly, to probe image features at different length scales and preserve dimensionality between all network layers, dilated convolutions \cite{yu2015multi} replace upscaling and downscaling operations typically found in convolutional neural networks. Like their non-dilated counterparts, convolutions of integer dilation $l$ consist of the the same square kernel, though the kernel's receptive field is expanded by spacing neighboring entries $(l-1)$ pixels apart; e.g. a two-dimensional $3 \times 3$ dilated convolution with $l=10$ has $9$ pixels between each of the vertically- and horizontally-adjacent entries in the kernel matrix, resulting in a receptive field of $21 \times 21$ pixels. Secondly, as depicted in a 3-layer MSDNet diagram in Fig.~\ref{fig:msdnet}, layers associated with different length scales are mixed together by densely connecting \emph{all} potential pairs of layers. This dense connectivity leads to several advantages, including maximum feature reusability, recovery of spatial information lost in the early layers, alleviation from the vanishing gradient problem \cite{ioffe2015batch} that plagues deep or stacked networks \cite{tong2017image}, and more robust model convergence and finer-grained predictions \cite{drozdzal2016importance}.

\begin{figure}
    \includegraphics[width=0.45\textwidth]{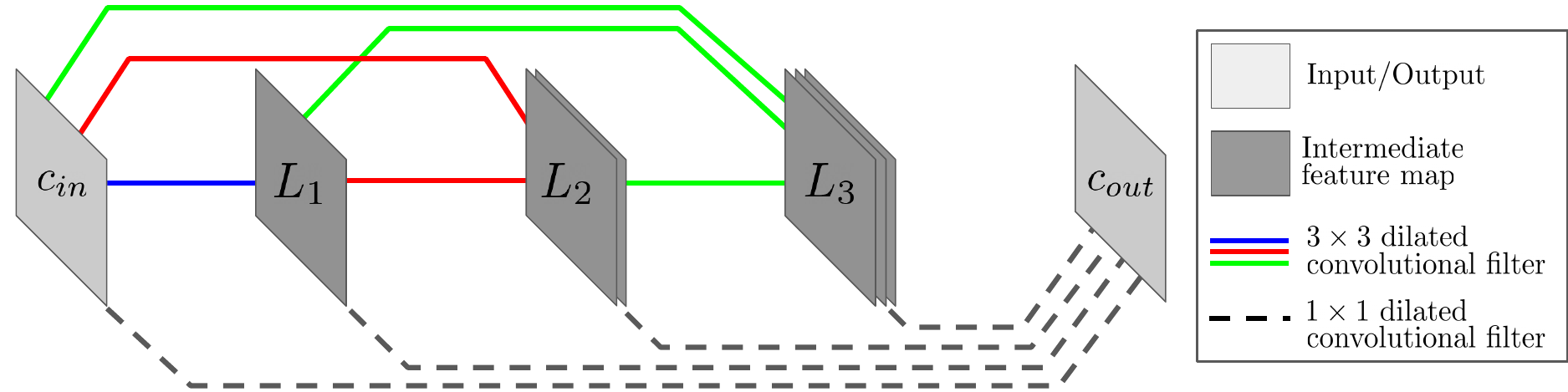}%
    \centering
        \caption{Schematic of a three-layer mixed-scale dense network (MSDNet) with $c_{in}$ and $c_{out}$ number of input and output channels. Blue, green, and red solid lines represent $3\times3$ dilated convolutions between all possible pairs of input and intermediate layers, with different dilations assigned to each color. Black dashed lines on the bottom connecting all input and intermediate layers to the output layer represent $1\times 1$ convolutional operators, amounting to a linear sum between individual pixels at each position among all non-output layers.}
        \label{fig:msdnet}
        
\end{figure}

MSDNets are distinct from encoder-decoder networks, as the \emph{same} set of operations are applied to all densely-connected layers: dilated $3\times 3$ convolutions with a layer-specific dilation, ReLU nonlinear activation, and batch normalization to expedite training. The final output layer is computed by replacing dilated convolutions with $1 \times 1$ non-dilated convolutions. These single pixel filters result in a linear combination with learnable weights of all pixels in a single position among previous layers whose weights. Overall, MSDNets have a much simpler architecture than the aforementioned U-Net design. As a result, DLSIA API requires only two main hyperparameters with which to govern the network architecture, namely:

\begin{enumerate}
    \item depth $d$: the number of network layers,
    \item maximum dilation $l_m$: the maximum integer dilation of the network; i.e. each layer $d_i$ is assigned integer dilation $i~\mod~l_m$,
    \begin{enumerate}
        \item custom dilations: alternatively, DLSIA users can manually assign specific dilations to each layer with a vector of length $d$; e.g. cycling through dilations of size $[1, 2, 4, 8, 16]$ ten times in a network with $d = 50$.
    \end{enumerate}
\end{enumerate}

\subsection{Sparse Mixed-scale Convolutional Neural Networks} \label{sect:smsnets}

Mixed-scale dense networks are designed to require a minimal number of parameters, yet the resulting networks may still be trimmed down using pruning approaches. For instance, results from the graph-based pruning method LEAN \cite{schoonhoven2020lean} demonstrate that large MSDNets can be reduced to 0.5\% of their original size without sacrificing significant performance. Given the high quality in performance of pruned networks in general \cite{blalock2020state, park2016faster, wang2021emerging}, it would be advantageous to be able to create \emph{pre-pruned} networks from scratch, aimed at producing networks that are as lean as possible with the lowest chances of overfitting.

In this communication, we aim to produce this type of network by using a stochastic approach that yields random networks with configurable complexity. These sparse mixed-scale networks (SMSNets), shown in Fig.~\ref{fig:SMSNet}, are stochastically configured, both topologically with varying random connections and morphologically with convolutions of different dilations assigned to each connection. This random nature of model architectures produces additional diversity among the models, making them suitable for ensemble methods \cite{ dietterich2000ensemble, ganaie2022ensemble}. Each SMSNet is produced using the following user-specified hyperparameters:

\begin{enumerate}
    \item $d$:  number of nodes between the input (I) node and the output (O) node.
    \item $k_{\min}$, $k_{\max}$: the global minimum and maximum number of edges per node. By default, these are set to $1$ and $(d+1)$, respectively. Adjustments are made on a node level based on their depth.
    \item $LL_\gamma$: the degree distribution parameter. The number of edges $n_j$ at node $j$ is a random number drawn from a distribution with density proportional to $\exp(-\gamma n_j)$, with $n_j \in [ {\min}(k_{\min},d-j), {\min}(k_{\max}, d-j)]$.
    \item $LL_\alpha$: skip-connection distribution parameter governing the probability for an edge to be assigned between node $i$ and node $j$, proportional to $\exp(-\alpha|i-j|)$.
    \item $P_{IL}$: the probability for an edge between input node $I$ and any of the intermediate hidden nodes $L$.
    \item $P_{LO}$: the probability for an edge between an intermediate hidden node $L$ and the output node $O$.
    \item $P_{IO}$: boolean variable that allows edges between all channels in input node $I$ and output node $O$.
\end{enumerate}

\noindent Below, in Sect.~\ref{sect:fibers}, we leverage predictions from an ensemble of several low parameter SMSNets, each with varied architectures generated stochastically and independently using the above hyperparameters available in DLSIA.

\begin{figure}
    \includegraphics[width=0.45\textwidth]{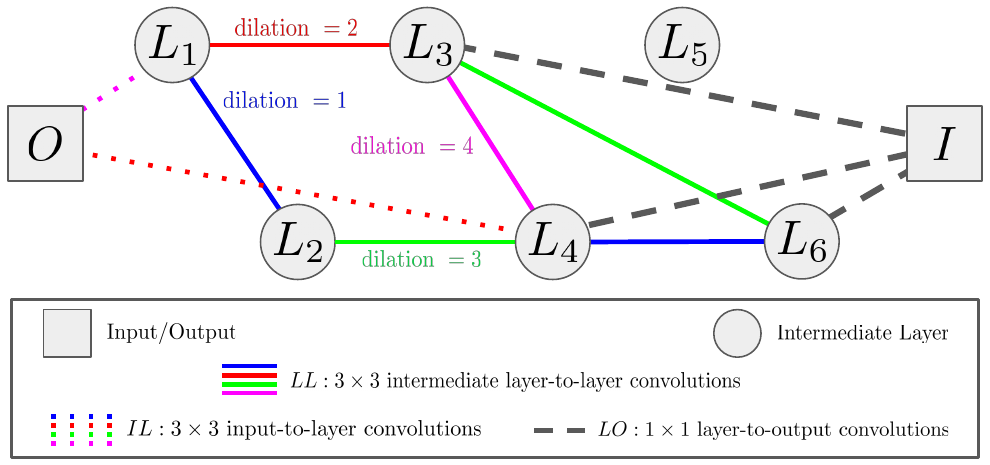}%
    \centering
        \caption{Schematic overview of a six-layer sparse mixed-scale network (SMSNet). Network nodes consist of the input data $I$, six intermediate (hidden) layers $L$, and output data $O$. All nodes/layers are sparsely connected via convolution filters, represented by dashed, dotted, and solid lines. For sake of simplicity, connections between input-to-output ($IO$) channels are not shown.}
        \label{fig:SMSNet}
        
\end{figure}

\section{Utility functions} \label{sect:utility}

In addition to custom CNN architectures, DLSIA offers a number of tools to assist in the end-to-end training process.

\begin{enumerate}

    \item \textbf{Training Scripts:} DLSIA offers comprehensive training scripts for effortlessly loading data and customizing training instances. Researchers can easily fine-tune a range of essential parameters, including optimizer selection, learning rate, learning schedulers, gradient clipping, early stopping, and automatic mixed precision. This flexibility ensures that users can tailor their training process to the unique demands of their scientific image analysis tasks, while efficiently optimizing model performance.
    
    \item \textbf{Custom Loss Functions:} In addition to standard classification loss functions such as the cross entropy provided by PyTorch, DLSIA provides a collection of custom loss functions designed to tackle specific challenges in scientific image analysis. The Dice loss \cite{sorensen1948method} is an alternative to the cross entropy loss that measures the overlap between predicted and ground truth masks. The Focal loss \cite{lin2017focal} aids in handling imbalanced datasets by prioritizing hard-to-classify samples during training. The Tversky loss \cite{tversky1977features} offers a fine-tuned balance between false positives and false negatives, granting users more control over the desired trade-offs during training.
    
    \item \textbf{Random Data Loaders:} 
    In PyTorch, random data splitters are often used for creating separate training, validation, and testing datasets from a larger dataset, a crucial step in training a robust machine learning model. These tools, such as the RandomSplit function, work by randomly assigning a certain proportion of the dataset to each subset. This ensures an unbiased distribution of data points, aiding in preventing overfitting and improving the generalization capability of the model. In essence, random data splitters provide a quick and efficient method to divide datasets, paving the way for effective model training and evaluation processes.
    
    While random data splitters in PyTorch excel in scenarios with large data volumes, their effectiveness can diminish in segmentation problems with a shortage of images. This is because they operate at the image level, meaning they can't split and shuffle small data sets effectively for robust training and testing. To overcome this limitation, the DLSIA introduces random data loaders that perform splitting at a more granular pixel level, creating randomized disjoint sets. This allows for more representative distributions of training and validation data, even in situations with limited images, leading to better model performance and generalizability.

    \item \textbf{Conformal Estimation Methods:} DLSIA offers conformal estimation methods \cite{angelopoulos2021gentle} enabling researchers to determine confidence intervals for their model predictions. By quantifying uncertainty in predictions, calibrated prediction sets with user-specified coverage are provided allowing one to make informed decisions in critical applications.
\end{enumerate}

\section{Applications using DLSIA} \label{sect:applications}

We use DLSIA in the following examples to build end-to-end deep learning workflows. Section \ref{sect:inpainting} uses MSDNets and tunable U-Nets for inpainting purposes. Here, network training was performed on a single $40$ GB capacity Nvidia A100 GPU. Additionally, Sects.~ \ref{sect:fibers} and \ref{sect:shapes} validating SMSNet ensembling and autoencoder latent space clustering were performed on a single $24$ GB memory capacity Nvidia RTX 3090 GPU, along with a 20-thread I9-10900X Intel Core CPU for loading, distributing, and receiving works calls to and from the GPU. All training was performed using the ADAM optimizer \cite{kingma2014adam}. 

\subsection{Inpainting X-ray Scattering images with U-Nets and MSDNets} \label{sect:inpainting}

Image inpainting is a restoration process that estimates the contents of missing regions within images and videos. Several machine learning (ML) approaches exist for inpainting \cite{inpainting_review, inapinting_review_comprehensive}, chief among them being competing dual-model generative adversarial networks (GANs) \cite{modified_gan, unsupervised_gan} and partial convolutional operators which augment traditional convolutional layers with adaptive kernel masking \cite{ part_conv_paper}. While inpainting results have recently gained popularity in non-scientific communities for it’s ability to blindly fill in pictures of heavily masked faces, inpainting in X-ray scattering sciences is limited to only a handful of previous studies which heavily exploit symmetry \cite{heal_scattering}. Since beamline scientists are currently using ML-based algorithms to process the large amount of data they collect \cite{ml_scattering}, it is of great importance to reconstruct the missing regions to avoid the introduction of distortion and bias to the post-processing ML analysis. 

Hence, DLSIA was employed to inpaint the missing pixel information in vertical and horizontal detector gaps in X-ray scattering datasets. In this 2022 study, published in \cite{chavez2022comparison}, the ground truth information exists for the missing horizontal gap data in which to train against, though missing gap data information is entirely nonexistent for the vertical bars. To alleviate this constraint, data augmentation was performed. Outlined in Fig.~\ref{fig:inpainting_overview}, this augmentation process artificially introduced vertical bar gaps in new positions which contained ground truth data behind them.

\begin{figure}
    \includegraphics[width=0.45\textwidth]{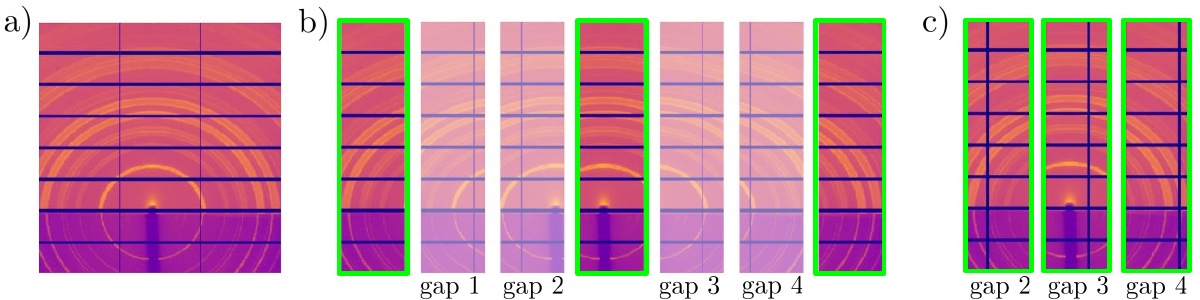}%
        \centering
        \caption{Inpainting data augmentation process to artificially present new vertical gaps with ground truth information behind them.  \textbf{a)} Input data is \textbf{b)} cropped into seven overlapping images, introducing new vertical gaps in one of four positions in the non-highlighted images. \textbf{c)} Highlighted images constitute the original input, but artificial gaps are randomly inserted in one of the four new gap positions.}
        \label{fig:inpainting_overview}
        
\end{figure}

Two distinct CNNs, a U-Net and an MSDNet, were used for leaning to inpaint the gaps. Once the data augmentation steps were complete, nearly 15,000 training images were used, of which three are shown in Fig.~\ref{fig:inpainting_overview}c. The $L_1$ loss metric, which gauges differences between gap predictions and ground truth, was chosen as the target function to minimize. The $L_2$ loss was also tested but resulted in more blurring, as is consistent with previous inpainting studies \cite{isola2017image}. Of several different network architectures tested, a depth-4 U-Net with ${\sim} 8.56$ million parameters and a 200-layer MSDNet with ${\sim} 0.18$ million parameters were the best performing networks, both achieving correlation coefficient scores of $> 0.998$ between predicted gaps and ground truth. These predictions are displayed in Fig.~\ref{fig:inpainting_results}.

\begin{figure}
    \includegraphics[width=0.45\textwidth]{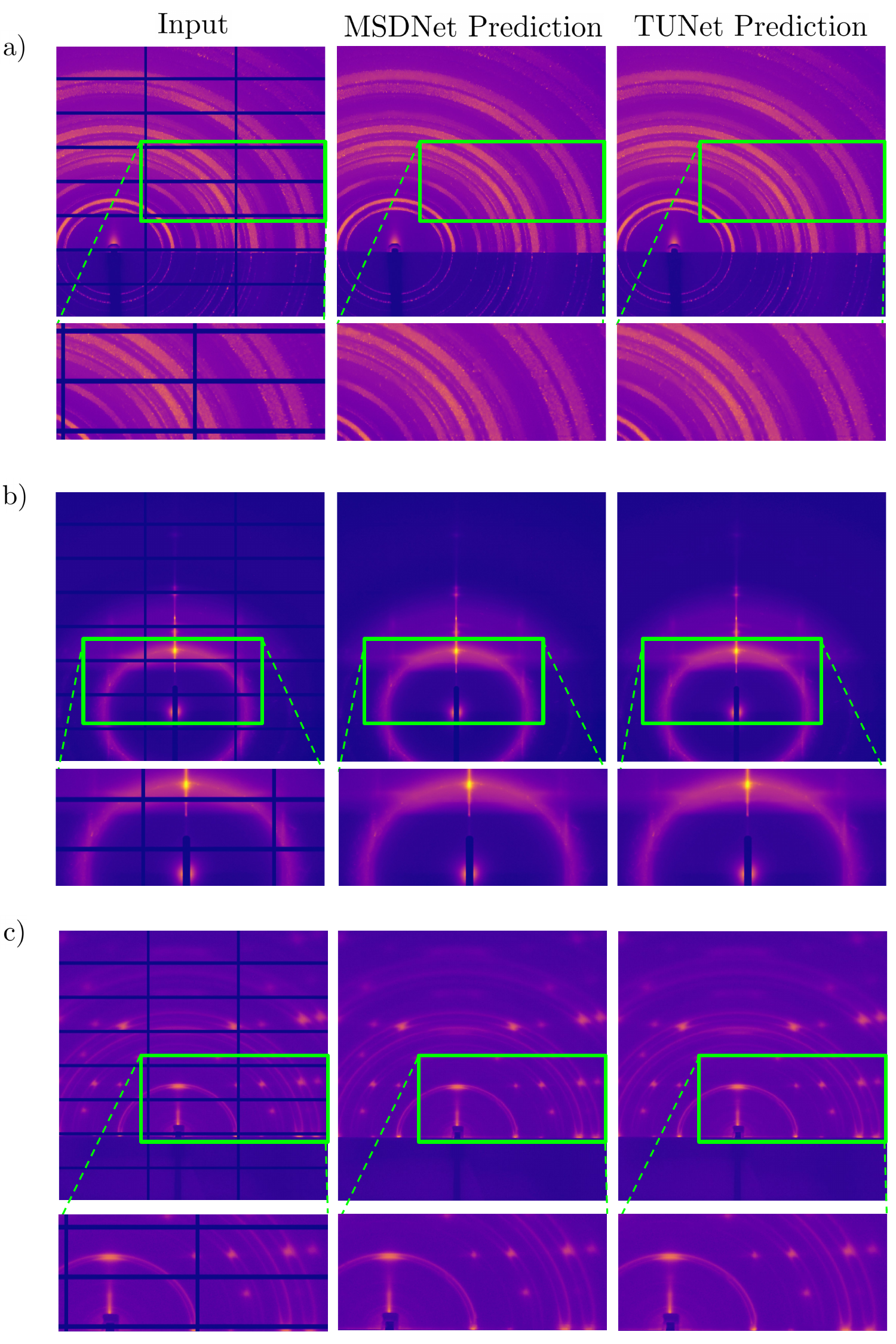}%
    \centering
    \caption{Inpainting of X-ray scattering vertical and horizontal detector gaps using U-Net and MSDNet.}        
    \label{fig:inpainting_results}

\end{figure}

\subsection{Detecting 3d fibers in X-ray tomographic reconstructions of concrete using SMSNet ensembling} \label{sect:fibers}

Fiber reinforcement in concrete plays a fundamental role in enhancing the material's properties, delivering increased tensile strength, superior shrinkage control, and enhanced flex-induced crack, blast, and fire resistances \cite{zollo1997fiber, naser2019fiber}. As concrete naturally excels under compression resistance but lags in tensile strength, fibers serve to augment this tensional weakness, ensuring the material can endure greater tensile stresses. Furthermore, fibers significantly contribute to the concrete's toughness and durability, providing heightened resistance to impact, abrasion damage, and shrinkage-related cracking. Simultaneously, the integral role of fibers in mitigating shrinkage throughout the curing process and the concrete's lifetime ensures overall enhanced longevity of the structure.

Understanding the structural distribution of fibers within the concrete matrix is pivotal for comprehending the properties of the composite material and consequently the design of better concrete mixtures. Fiber distribution, orientation, and density greatly impact the overall performance of the concrete, influencing its strength, ductility, and fracture resistance. This characterization can be achieved through techniques such as X-ray tomography. However, concrete is a complex mixture comprising various components, such as cement, aggregates, and fibers. Isolating and identifying fibers within the voluminous and intricate 3D data obtained from X-ray tomography is not a straightforward task. 

In this context, we used a publicly available dataset and performed manual binary segmentation of select fiber sections to curate a ground truth for supervised learning. Manual segmentation using Napari software \cite{sofroniew2022napari} consisted of the sparse and incomplete hand-annotation of only 6 fibers, consisting of ${\sim} 245,000$ labeled pixels with a $10 : 2$ background-to-foreground ratio. Hand-annotations focused on a diverse representation of cases and contrasts among the fibers is displayed in Fig.~ \ref{fig:fiberLengths}a. This selection was restricted to a few locations with the focus of balancing accuracy -- particularly when labeling the border between classes -- and overall speed of annotation to maintain a manageable workload.

\begin{figure}[h]
  \centering
	\includegraphics[width=0.45\textwidth]{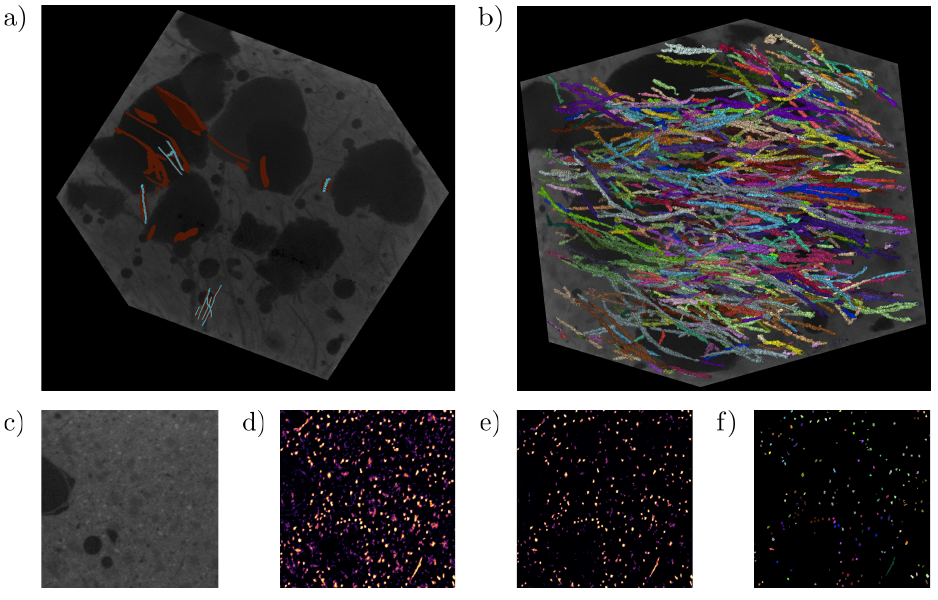}
  \caption{Ensemble network predictions of fibers in concrete.  ~\textbf{a.} Sparse binary labeling of target fibers (cyan) and background (brown). ~\textbf{b.} Aggregated network predictions. ~\textbf{c.} Cross sectional slice of raw training data. ~\textbf{d.} Probability map of aggregated network predictions. ~\textbf{e.} Probability map with standard deviations subtracted. ~\textbf{f.} Cross sectional view of instance segmented fibers derived from \textbf{e}.}
  \label{fig:fibers}
  
\end{figure}

The prepared data was then subjected to an ensemble of five DLSIA-instantiated sparse mixed-scale networks (SMSNets), each with a different stochastically generated architecture and approximately 45,000 parameters. The multi-network mean prediction probabilities are displayed in \ref{fig:fiberLengths}d. However, we choose to leverage the multi-network standard deviation and keep only those pixel predictions whose probability remains over $50\%$ after subtracting a single standard deviation, pictured in \ref{fig:fiberLengths}e. A subsequent analysis using the external Python package \texttt{cc3d} \cite{silversmith2021cc3d} involved 3d instance segmentation using  a decision tree-augmented 3d variant of connected components \cite{wu2005two}. Additionally, \texttt{cc3d} allowed for the removal of small connected components -- a so-called "dusting" -- below some user-defined threshold. Both a histogram of the end-to-end length of the instance segmented fibers and a Hammer projection \cite{tobler1964some} of the surface an origin-centered 30-pixel sphere of the autocorrelation function of the segmented labels -- essentially measuring the directional distribution of the segmented fibers -- is shown in Fig.~\ref{fig:fiberLengths}, providing critical insights into the morphology and organization of the segmented fibers that can be used to understand, predict, or design properties of fiber-reinforced concrete.

\begin{figure}[h]
  \centering
  \includegraphics[width=0.48\textwidth]{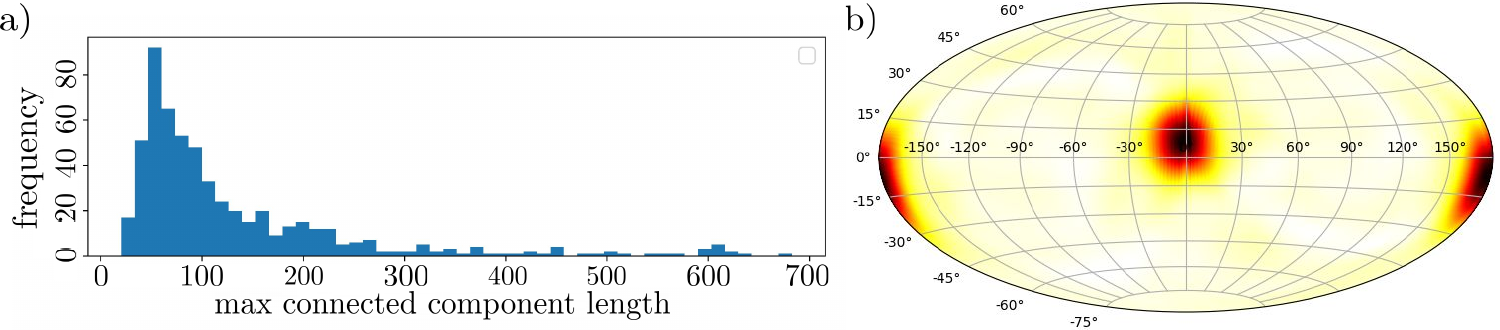}
  \caption{Summary statistics of fiber segmentation predictions. Displayed are ~\textbf{a.} histogram plot of fiber lengths and ~\textbf{b.} equal-area Hammer projection of the autocorrelation function of the 3D segmentation results at a radius of 30 pixels from the origin showing a general anisotropic distribution of the direction of the fibers.}
  \label{fig:fiberLengths}
\end{figure}



\subsection{Shape clustering via autoencoder latent space} \label{sect:shapes}

We present the results of our clustering approach on the highly compressed autoencoder latent space using synthetic data consisting of $64 \times 64$ tiles, each containing one of four random shapes (circle, triangle, rectangle, and annulus) that are randomly sized and rotated by a random degree around their centers. We apply a 4-layer 16-base channel autoencoder that bottlenecks to a $16 \times 1$ sized latent space (or feature space) to reconstruct the input data, optimized on the mean square error loss. To assess the quality of our model reconstruction, we find the Pearson cross-correlation scored against the original images which yielded an impressive score of approximately 0.98.

Once the model is sufficiently trained, we pass new images through the trained autoencoder to obtain their $16 \times 1$ latent space representation. To visualize and analyze the clustering behavior, we further compress the latent space down to two real numbers using U-Map \cite{mcinnes2018umap}, allowing us to generate meaningful scatter plots in Cartesian coordinates. As illustrated in Fig.~\ref{fig:autoencoderlatentspace}, our approach exhibits clear, distinct clustering results between each of the four shapes. Moreover, the approach handles the variations in shape orientation and size remarkably well, with clear transitions between each shapes' size and orientation within each cluster.

\begin{figure}
    \includegraphics[width=0.45\textwidth]{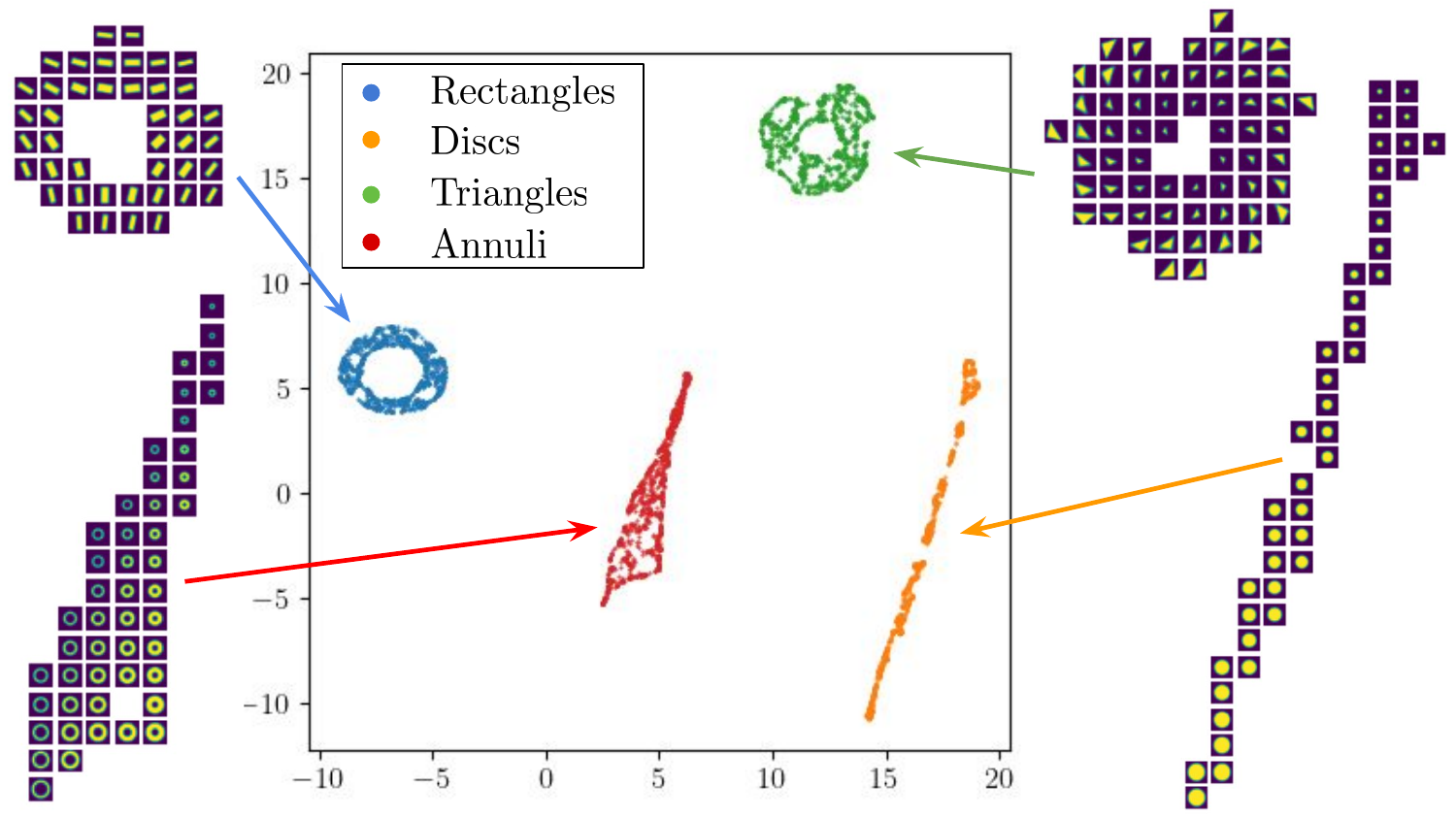}
        \centering
        \caption{Autoencoder latent space representation, further compressed by U-Map, of randomly sized and oriented shapes.}
        \label{fig:autoencoderlatentspace}
        
\end{figure}

\section{Discussion and Conclusions} \label{sect:discussion}

We introduce DLSIA (Deep Learning for Scientific Image Analysis), a Python-based deep learning convolutional neural network library aimed at bringing a new level of user-customizability to researchers and their image analysis tasks. Offering simplified network construction, multiple proven network architectures, and an array of tunable training parameters, DLSIA provides a versatile platform allowing users to explore diverse network settings. DLSIA-instantiated networks and workflows were validated through three separate applications: 1) semantic segmentation of fibers in X-ray tomographic reconstruction of concrete data using an ensemble of sparse mixed-scale networks (SMSNets), 2) inpainting of missing gap information in X-ray scattering data using U-Nets and mixed-scale dense networks (MSDNets), and 3) investigation into clustering autoencoder latent space on synthetic shapes data.

\section*{Acknowledgments}
The above algorithms are implemented in a set of python3 routines, and are available upon request. Additionally, DLSIA modules for custom MSDNet, autoencoder, and U-Net instantiation for segmentation purposes are available within MLExchange \cite{zhao2022mlexchange, hao2023deploying}. MLExchange is a DOE-funded, web-based collaborative platform offering accessible machine learning tools for beamline scientists, including a custom API for managing jobs and workflows, adaptive GUI components for labeling data and customizing models, and trained model registration and distillation \cite{hexemer2021mlexchange}.

We gratefully acknowledge the support of this work by the Laboratory Directed Research and Development Program of Lawrence Berkeley National Laboratory under U.S. Department of Energy Contract No. DE-AC02-05CH11231. Further support originated from the Center for Advanced Mathematics in Energy Research Applications funded via the Advanced Scientific Computing Research and the Basic Energy Sciences programs, which are supported by the Office of Science of the US Department of Energy (DOE) under Contract DE-AC02-05CH11231, and from the National Institute Of General Medical Sciences of the National Institutes of Health (NIH) under Award 5R21GM129649-02. The content of this article is solely the responsibility of the authors and does not necessarily represent the official views of the NIH. The inpainting study was performed and partially supported by the US DOE, Office of Science, Office of Basic Energy Sciences Data, Artificial Intelligence and Machine Learning at the DOE Scientific User Facilities program under award No. 107514.

\bibliographystyle{IEEEtran}
\bibliography{IEEEabrv, bibliography}

\newpage

\section*{Supplementary Information} \label{sect:supplementary}

The DLSIA library contains many subroutines and functionalities not listed or mentioned in the text above. Table \ref{table:scripts} references a bulk of the DLSIA modules available for use. For full documentation and listing of modules, please see \url{https://dlsia.readthedocs.io/en/latest/}.

\begin{table}[h]
    \centering
    \renewcommand{\arraystretch}{1.8} 
    
    \begin{tabular}{p{0.24\linewidth} | p{0.66\linewidth}}
        \hline
         Script/Module & Description \\ \hline
        
         baggins.py & Contains ensembling-based modules for bagging models.\\ 
        
         \makecell[l]{ conformalize\_\\segmentation.py} & Used to perform conformal estimation on a set of model predictions.\\ 

         
         custom\_losses.py & Contains an array of popular loss functions suitable for image segmentation.\\ 
        
         \makecell[l]{draw\_sparse\_\\network.py} & Visualizes the topology and layout of individual SMSNets.\\ 
        
         helpers.py & Contains several minor utility functions, including functions for retrieving the current computing device, counting model parameters and convolutional filters, and initiating PyTorch DataLoader classes.\\ 
        
         \makecell[l]{latent\_space\_\\viewer.py}  & Visualizes images in autoencoder latent space upon a single instance of U-Map, as viewed in Fig.~\ref{fig:autoencoderlatentspace}.\\
        
         msae.py & Creates autoencoder networks. Mixed-scale functionality is forthcoming.\\ 
        
         msdnet.py & Creates mixed-scale dense networks (MSDNets).\\ 

         plots.py & Contains suite of plotting tools for model segmentation, regression, and aggregation. \\
        
         random\_shapes.py & Generates random circles, rectangles, triangles, and annuli used in Sect.~\ref{sect:shapes} with random size, orientation, and user-defined Gaussian noise.\\ 
        
         \makecell[l]{randomized\_data\_\\loader.py} & Returns input data into random partition of training and testing data.\\ 

         scale\_up\_down.py & Contains modules for data resizing used in TUNets, U-Nets, and autoencoders. \\
         
         \makecell[l]{segmentation\_\\metrics.py} & Computes F1 scores for evaluating quality of model segmentation performace. \\
        
         smsnet.py & Creates random, sparse mixed-scale networks (SMSNets) for 2D data.\\
         
         smsnet3d.py & Creates random, sparse mixed-scale networks (SMSNets) for 3D data.\\
         
         train\_scripts.py & Contains end-to-end model training procedures and evaluation metrics for segmentation and regression problems.\\ 
        
         tunet.py & Creates custom, tunable U-Nets.\\ 
        
         tunet3plus.py & Creates U-Net3+, a new variant of the classic
            U-Net featuring dense skip connections aggregating features from all network layers.\\
        
        \hline
    \end{tabular}
    \vspace*{0mm}
    \caption{DLSIA utility modules and functions}
    \label{table:scripts}
\end{table}

\end{document}